\documentclass[sigconf]{acmart}

\AtBeginDocument{
  }

\setcopyright{acmlicensed}
\copyrightyear{2018}
\acmYear{2018}
\acmDOI{XXXXXXX.XXXXXXX}
\acmConference[Conference acronym 'XX]{Make sure to enter the correct
  conference title from your rights confirmation email}{June 03--05,
  2018}{Woodstock, NY}

\usepackage{wasysym}

\usepackage{algorithm}
\usepackage{algorithmic}
\usepackage{tcolorbox}
\usepackage{fontawesome5}
\usepackage{pifont}
\newcommand{\cmark}{\ding{51}}
\newcommand{\xmark}{\ding{55}}

\usepackage{tcolorbox}
\tcbuselibrary{skins}
\newtcolorbox{PromptTextBox}[1]{
  colback=gray!5,
  colframe=black,
  boxrule=0.5pt,
  arc=2mm,
  width=\linewidth,
  title={#1},
  fonttitle=\bfseries,
  enhanced,
  left=2mm,
  right=2mm,
  top=1mm,
  bottom=1mm,
  boxsep=1mm
}

\usepackage{enumitem}
\setlist[itemize]{leftmargin=*}
\usepackage{CJKutf8}

\usepackage{tabularx}
\usepackage{xcolor}
\usepackage{colortbl}

\definecolor{googleblue}{RGB}{66,133,244}
\definecolor{googlered}{RGB}{219,68,55}
\definecolor{googlegreen}{RGB}{15,157,88}
\definecolor{googlepurple}{RGB}{138,43,226}
\definecolor{lightred}{RGB}{255, 220, 219}
\definecolor{lightblue}{RGB}{204, 243, 255}
\definecolor{lightgreen}{RGB}{200, 247, 200}
\definecolor{lightpurple}{RGB}{230,230,250}
\definecolor{lightyellow}{RGB}{242, 232, 99}
\definecolor{lighterblue}{RGB}{197, 220, 255}
\definecolor{lighterred}{RGB}{253, 249, 205}
\definecolor{lightyellow}{RGB}{207, 161, 13}
\definecolor{darkpurple}{RGB}{218, 210, 250}
\definecolor{darkred}{RGB}{255,198,196}
\definecolor{darkblue}{RGB}{172, 233, 252}

\usepackage{amsmath}
\usepackage{comment}
\usepackage{graphicx}
\usepackage{multirow}
\usepackage{booktabs}
\usepackage{caption}
\usepackage{array}
\usepackage{hyperref}

\usepackage{pifont}

\newcolumntype{H}{>{\setbox0=\hbox\bgroup}c<{\egroup}@{}}

\definecolor{thedarkblue}{RGB}{0,0,120}
\definecolor{mygreen}{RGB}{0,100,0}
\definecolor{mydarkblue}{rgb}{0,0.08,0.45}
\definecolor{darkblue}{rgb}{0,0.08,180}
\definecolor{orange}{HTML}{E6550D}
\colorlet{TufteRed}{red!80!black}

\definecolor{theblue}{RGB}{0,0,180}
\colorlet{thered}{TufteRed}

\providecommand{\eat}[1]{\ignorespaces}
\providecommand{\addressed}[1]{\ignorespaces}
\providecommand{\todoeat}[1]{\ignorespaces}
\providecommand{\supp}[1]{\ignorespaces}

\DeclareMathOperator{\hugeE}{\mbox{\huge\raise-0.3ex\hbox{E}}}
\DeclareMathOperator{\p}{\mathbb{P}}
\DeclareMathOperator{\hugep}{\mbox{\huge\raise-0.3ex\hbox{$\p$}}}

\providecommand{\cmark}[0]{\checkmark}

\begin{document}
\title[TSDS-Toolbox: A Toolbox for Measuring Time-Series Dataset Similarity]
{TSDS-Toolbox: A Toolbox for Measuring\\
Time-Series Dataset Similarity}

\author{Yen-Ku Liu}
\email{yenkuliu.cs15@nycu.edu.tw}
\affiliation{
  \institution{National Yang Ming Chiao Tung University}
  \city{Hsinchu}
  \country{Taiwan}
}

\author{Hongjie Chen}
\email{hongjie.chen@dolby.com}
\orcid{0000-0002-8755-2099}
\affiliation{
  \institution{Dolby Laboratories}
  \city{Atlanta}
  \state{Georgia}
  \country{USA}
}

\author{Ryan A. Rossi}
\email{ryrossi@adobe.com}
\affiliation{
  \institution{Adobe Research}
  \city{San Jose}
  \state{California}
  \country{USA}
}

\author{Franck Dernoncourt}
\email{dernonco@adobe.com}
\orcid{0000-0002-1119-1346}
\affiliation{
  \institution{Adobe Research}
  \city{Seattle}
  \state{Washington}
  \country{USA}
}

\renewcommand{\shortauthors}{Liu et al.}

\begin{abstract}
The rapid advancement of artificial intelligence (AI) has significantly accelerated research in time-series analysis, particularly in forecasting, classification, and generation tasks.
Recent models, especially foundation models, benefit from time-series dataset similarity due to its significant role in source dataset selection for fine-tuning.
However, many existing implementations for benchmarking time-series dataset similarity methods are fragmented and difficult to extend.
To address this, we present a unified framework, the Time-Series Dataset Similarity Toolbox (\textbf{TSDS-Toolbox}).
Our work enables
(1) systematic and reproducible comparisons of time-series dataset similarity methods; 
(2) flexible extensibility for users to add customized datasets, similarity methods, and downstream time-series tasks; and
(3) consistent evaluation of both dataset-level and series-level similarity methods through integrated time-series dataset reducers.
The effectiveness of TSDS-Toolbox is validated through comprehensive experiments under diverse experimental settings.
Our toolbox is publicly available.
\footnote{
Our code:~\url{https://github.com/yenkuliu/TSDS-Toolbox}
}
\end{abstract}


\maketitle

\section{Introduction}
Time series are pervasive in domains such as finance, healthcare, speech processing, and climate science, where they provide a natural representation of temporally evolving systems~\cite{zhang2017stock, che2018recurrent, baevski2020wav2vec, nguyen2023climax, wu2022timesnet}.
As time-series research has expanded, dedicated resources have been developed for classification~\cite{bagnall2018uea}, anomaly detection~\cite{paparrizos2022tsb}, and foundation-model development~\cite{ansari2024chronos, das2023decoder, woo2024unified}.
This makes relevant source-dataset retrieval and selection important for downstream time-series tasks, particularly when models rely on transfer, adaptation, or external data reuse~\cite{han2025retrieval, ning2026ts, jin2024time, liu2024unitime}.

Quantifying similarity between time-series datasets supports several core workflows, including source-dataset selection, dataset visualization and inspection, benchmark curation, and analysis of foundation-model generalization~\cite{sun2025quantifying, ehrig2024impact, zhang2026unified, yao2025estimating}.
In transfer-learning and adaptation settings, a source dataset that is closer to the target domain is more likely to provide useful fine-tuning signals, while the resulting similarity structure can also help characterize cross-domain generalization behavior.

Existing time-series benchmarks and toolboxes mainly support classification~\cite{harutyunyan2019multitask}, forecasting~\cite{wang2026deep, loning2019sktime}, and general time-series model development~\cite{loning2019sktime}, while similarity toolboxes primarily focus on individual time series rather than dataset-level comparison~\cite{qiu2024tfb, tavenard2020tslearn}.
Consequently, time-series dataset similarity metrics are often evaluated under inconsistent experimental settings~\cite{sun2025quantifying, zhang2023similarity}, requiring researchers to repeatedly integrate metrics, standardize dataset formats, and build downstream evaluation pipelines from scratch for each study.
To address these limitations, we propose Time-Series Dataset Similarity Toolbox (\textbf{TSDS-Toolbox}), a unified and extensible framework for benchmarking time-series dataset similarity methods and evaluating their downstream utility.
Our main contributions are:
\begin{enumerate}
    \item A unified benchmarking interface enables fair comparison of time-series dataset similarity methods under consistent experimental settings.
    \item Standardized downstream evaluation pipelines assess the practical utility of dataset similarity for time-series tasks.
    \item A modular and configuration-driven design supports reproducible experiments and future extensions.
\end{enumerate}

\begin{figure*}[t]
    \centering
    \includegraphics[width=1\textwidth]{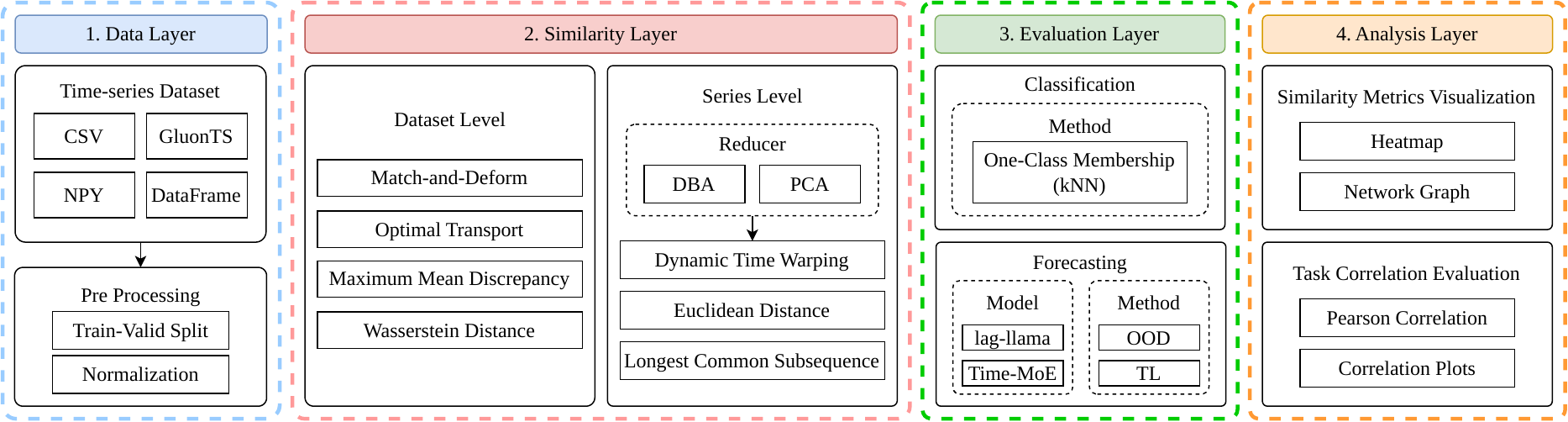}
    \caption{Overview of the TSDS-Toolbox architecture and its modules. 
    }
    \label{fig:architecture}
\end{figure*}

\section{Related Work}
Time-series similarity is commonly studied through sequence-level distances such as Dynamic Time Warping (DTW) and Longest Common Subsequence (LCSS)~\cite{sakoe1978dynamic, cuturi2017soft, vlachos2002discovering}, which compare individual sequences rather than datasets.
For practical dataset-level comparison, these distances can be combined with reducers such as DTW Barycenter Averaging (DBA)~\cite{petitjean2011global} and Principal Component Analysis (PCA)~\cite{pearson1901liii} to obtain compact representative sequences before computing similarity~\cite{fawaz2018transfer}.
A complementary direction formulates datasets as empirical distributions~\cite{alvarez2020geometric, heusel2017gans}, where methods such as Optimal Transport (OT) measure discrepancy through minimum-cost transport plans~\cite{gabriel2019computational}, and Maximum Mean Discrepancy (MMD) compares distributions using kernel mean embeddings~\cite{gretton2012kernel}.
Wasserstein Distance (WSD) has also been applied to time-series dataset similarity by modeling each dataset as an empirical multivariate Gaussian and relating the resulting distance to out-of-distribution and transfer-learning inference loss~\cite{chen2025measuring}.
Beyond these distributional metrics, Match-and-Deform (MAD) combines optimal transport with DTW-based temporal alignment, making it relevant for capturing both distribution shift and temporal deformation across time-series datasets~\cite{painblanc2023match}.

Existing time-series benchmarks and toolboxes support model development and evaluation for forecasting, classification, generation, and representation learning~\cite{alexandrov2020gluonts, herzen2022darts, loning2019sktime}, but they are not designed for systematic benchmarking of time-series dataset similarity metrics.

\section{Preliminaries}
We formalize the time-series dataset similarity problem before describing the toolbox design.
Let $X=\{x_i\}_{i=1}^{N_X}$ and $Y=\{y_j\}_{j=1}^{N_Y}$ denote two time-series datasets, where each sequence $x_i=(x_{i,1},\ldots,x_{i,T_i})$ and $y_j=(y_{j,1},\ldots,y_{j,T_j})$ may have different lengths, with observations in $\mathbb{R}^d$.
Time-series dataset similarity aims to define a dissimilarity function $s(X,Y)$ that quantifies the difference between two datasets, where smaller values indicate higher similarity.
We consider two classes of methods: dataset-level metrics, which directly compute $s_m(X,Y)=m(X,Y)$, and reducer-based series-level metrics, which first map each dataset to a representative sequence using a reducer $R$ and then compute $s_{R,d}(X,Y)=d(R(X),R(Y))$, where $d$ denotes a series-level distance.

\section{Time-series Dataset Similarity Toolbox}
This section presents the design of the Time-Series Dataset Similarity Toolbox (\textbf{TSDS-Toolbox}).
We first describe the overall pipeline supported by the toolbox (Sec.~\ref{sec:pipeline}).
We then introduce its four core layers: the Data Layer (Sec.~\ref{sec:data}), Similarity Layer (Sec.~\ref{sec:similarity}), Evaluation Layer (Sec.~\ref{sec:evaluation}), and Analysis Layer (Sec.~\ref{sec:analysis}).

\subsection{Overall Pipeline}
\label{sec:pipeline}

TSDS-Toolbox provides a configuration-driven pipeline for computing and evaluating time-series dataset similarity.
The pipeline supports two execution modes: a similarity-only mode for computing and visualizing pairwise dataset-distance matrices, and a similarity-evaluation mode for additionally running downstream tasks and producing task-performance matrices.
Users specify dataset formats, preprocessing settings, similarity methods, reducers, evaluation tasks, foundation-model adapters, and output options through YAML files, and can extend the toolbox with new dataset loaders, similarity metrics, reducers, evaluation tasks, model adapters, and analysis outputs through the corresponding toolbox layers.
Figure~\ref{fig:architecture} provides an overview of the layer organization and the modules within each layer in TSDS-Toolbox.

\begin{table*}[t]
\centering
\caption{Supported similarity methods and their constraints in TSDS-Toolbox. 
\cmark/\xmark denote support/non-support under the default setting; $N$ and $T$ denote the number of time series and sequence length. 
$^{*}$: multivariate support after flattening; $^{\dagger}$: Euclidean distance requires equal length by default but supports truncate and resample policies.}
\label{tab:supported_similarity_methods}
\resizebox{\textwidth}{!}{
\begin{tabular}{lccccccc}
\toprule
\textbf{Metric} & \textbf{Level} & \textbf{Reducer} & \textbf{Equal Length} & \textbf{Equal Feature Dim.} & \textbf{Multivariate} & \textbf{Different $N$} & \textbf{Different $T$} \\
\midrule
Wasserstein Distance & Dataset & \xmark & \cmark & \cmark & \xmark & \cmark & \xmark \\
Maximum Mean Discrepancy & Dataset & \xmark & \xmark & \cmark & \cmark$^{*}$ & \cmark & \xmark \\
Optimal Transport & Dataset & \xmark & \xmark & \cmark & \cmark$^{*}$ & \cmark & \xmark \\
Match-and-Deform & Dataset & \xmark & \xmark & \xmark & \cmark & \cmark & \cmark \\
Dynamic Time Warping & Series & \cmark & \xmark & \xmark & \cmark & N/A & \cmark \\
Euclidean Distance & Series & \cmark & \cmark$^{\dagger}$ & \cmark$^{\dagger}$ & \cmark & N/A & \xmark$^{\dagger}$ \\
Longest Common Subsequence & Series & \cmark & \xmark & \xmark & \cmark & N/A & \cmark \\
\bottomrule
\end{tabular}
}
\end{table*}

\subsection{Data Layer}
\label{sec:data}

As shown in Fig.~\ref{fig:architecture}-(1), the Data Layer loads time-series datasets from heterogeneous formats, including GluonTS~\cite{alexandrov2020gluonts}, CSV, NPY, and pandas DataFrames, and converts them into a unified representation with values and metadata for later stages.
It further applies user-specified preprocessing for both similarity computation and downstream evaluation, including sampling, fixed-length window extraction, reshaping, train-validation splitting, and optional normalization in a consistent manner.

\subsection{Similarity Layer}
\label{sec:similarity}

As shown in Fig.~\ref{fig:architecture}-(2), the Similarity Layer computes pairwise distance matrices over the preprocessed datasets.
It supports dataset-level metrics that compare two datasets directly, as well as series-level metrics that compare representative sequences produced by dataset reducers.
This unified interface allows both classes of similarity methods to be evaluated within the same pipeline, as summarized in Table~\ref{tab:supported_similarity_methods}.

\noindent\textbf{\textit{Dataset-Level Similarity Metrics.}}
Dataset-level metrics directly compute a dissimilarity score between two time-series datasets.
TSDS-Toolbox implements four representative metrics: \textit{Wasserstein Distance}, \textit{Maximum Mean Discrepancy}, \textit{Optimal Transport}, and \textit{Match-and-Deform}~\cite{chen2025measuring, gretton2012kernel, alvarez2020geometric, painblanc2023match}.
For fixed-dimensional metrics, each preprocessed time-series window is flattened so that datasets \(X\) and \(Y\) can be compared as empirical sample sets.

\textbf{Wasserstein Distance (WSD)}~\cite{chen2025measuring} is implemented as a Fréchet-style distance between Gaussian approximations of two empirical distributions, using the sample mean \(\mu_X\) and sample covariance \(\Sigma_X\):
\[
s_{\mathrm{WSD}}(X,Y)
=
\left(
\|\mu_X-\mu_Y\|_2^2
+
\mathrm{Tr}\left(
\Sigma_X+\Sigma_Y
-
2(\Sigma_X\Sigma_Y)^{1/2}
\right)
\right)^{1/2}.
\]

\textbf{Maximum Mean Discrepancy (MMD)}~\cite{gretton2012kernel} measures distributional discrepancy through a kernel function, with the default biased estimator:
\[
\begin{aligned}
s_{\mathrm{MMD}}(X,Y)
&=
\left[
\max\left(
\frac{1}{N_X^2}\sum_{i,i'} k(x_i,x_{i'})
+
\frac{1}{N_Y^2}\sum_{j,j'} k(y_j,y_{j'})
\right.\right. \\
&\qquad\left.\left.
-
\frac{2}{N_XN_Y}\sum_{i,j} k(x_i,y_j),
0
\right)
\right]^{1/2}.
\end{aligned}
\]

\textbf{Optimal Transport (OT)}~\cite{alvarez2020geometric} finds a minimum-cost transport plan between samples, where \(C_{ij}\) is the ground cost and \(a,b\) are normalized sample weights:
\[
s_{\mathrm{OT}}(X,Y)
=
\min_{\Pi\in U(a,b)}
\sum_{i=1}^{N_X}\sum_{j=1}^{N_Y}
\Pi_{ij}C_{ij}.
\]
Here, \(U(a,b)\) denotes the set of nonnegative transport plans with marginals \(a\) and \(b\), which are uniform by default. The toolbox supports exact Earth Mover's Distance and Sinkhorn approximation.

\textbf{Match-and-Deform (MAD)}~\cite{painblanc2023match} combines sample-level matching with temporal alignment by alternating between transport-plan estimation and DTW-based alignment updates. In the implementation, the returned score is
\[
s_{\mathrm{MAD}}(X,Y)=
\sum_{i=1}^{N_X}\sum_{j=1}^{N_Y}
\widehat{\Pi}_{ij}
\left(
\alpha \frac{1}{\bar{T}}
\sum_{(u,v)\in \widehat{W}_{c_i}}
\|x_{i,u}-y_{j,v}\|_2^2
+\beta B_{ij}
\right).
\]
Here, \(\widehat{W}_{c_i}\) is the final DTW path, \(\widehat{\Pi}\) is the final transport plan, \(B_{ij}\) is an optional pairwise cost, and \(\bar{T}\) is the normalization length. Defaults are \(\alpha=1\), \(\beta=0\), uniform weights, and normalized costs.

\noindent\textbf{\textit{Series-Level Similarity Metrics.}}

TSDS-Toolbox supports two reducers: \textbf{DTW Barycenter Averaging (DBA)} and \textbf{Principal Component Analysis (PCA)}~\cite{petitjean2011global, pearson1901liii}.
DBA averages time series under DTW-based alignment, while PCA constructs a representative sequence from dominant principal components of flattened time-series samples.

After reduction, the toolbox supports three series-level distances: \textit{Dynamic Time Warping}, \textit{Euclidean Distance}, and \textit{Longest Common Subsequence}~\cite{sakoe1978dynamic, ding2008querying, faloutsos1994fast, vlachos2002discovering}.
\textbf{Dynamic Time Warping (DTW)}~\cite{sakoe1978dynamic} finds a minimum-cost alignment using squared Euclidean local costs and returns the square root of the accumulated cost; \textbf{Euclidean Distance (ED)}~\cite{ding2008querying} computes the standard pointwise \(\ell_2\) distance with optional truncation or resampling; and \textbf{Longest Common Subsequence (LCSS)}~\cite{vlachos2002discovering} returns the normalized distance \(1-\mathrm{LCSS}(r_X,r_Y)/\min(|r_X|,|r_Y|)\) by default, where matches are defined by a Euclidean threshold and an optional temporal window.

\subsection{Evaluation Layer}
\label{sec:evaluation}
As shown in Fig.~\ref{fig:architecture}-(3), the Evaluation Layer assesses whether dataset similarity scores are informative for downstream time-series tasks.
It supports two evaluation settings, \textbf{out-of-distribution (OOD)} evaluation and \textbf{transfer-learning (TL)} evaluation, across two downstream task types, classification and forecasting.

\noindent\textbf{\textit{Classification Evaluation.}}
TSDS-Toolbox evaluates classification through a one-class out-of-distribution criterion, using the source dataset to define the reference distribution and testing target samples with a \(k\)-nearest-neighbor (\(k\)-NN) membership distance~\cite{cover1967nearest}.
Samples are accepted as source-like when their membership distance is below a source-calibrated quantile threshold \(\tau_X\).
The classification dissimilarity score is then defined as $E_{\mathrm{cls}}(X,Y) = 1-\mathrm{MembershipRate}(Y \mid X)$.
Lower values indicate that more target samples are accepted by the source-calibrated rule.

\noindent\textbf{\textit{Forecasting Evaluation.}}
TSDS-Toolbox uses model adapters to integrate foundation models such as Lag-Llama~\cite{rasul2023lag} and Time-MoE~\cite{shi2025time} into a unified fine-tuning and prediction pipeline.
In the out-of-distribution setting, a model fine-tuned on the source dataset is evaluated on each target dataset, and the Mean Squared Error (MSE) of its forecasts is recorded as the source-target task-performance score, \(E_{X,Y}=\mathrm{MSE}(f_X,Y_{\mathrm{inference}})\), where \(f_X\) denotes the model trained on source dataset \(X\) and MSE averages squared forecast errors over each prediction horizon and then over target inference samples.
In the transfer-learning setting, the model is first fine-tuned on the source dataset, then further adapted using a small target-reference split, and finally evaluated on the target-inference split to assess whether dataset similarity reflects transferability in realistic downstream adaptation scenarios.

\begin{table}[t]
\centering
\small
\setlength{\tabcolsep}{3pt}
\caption{Correlation between dataset similarity and downstream task loss. Cls. denotes classification; OOD-Lag and OOD-TMoE denote forecasting OOD evaluation with Lag-Llama and Time-MoE, respectively; TL-TMoE denotes forecasting transfer learning with Time-MoE. The symbol ``--'' indicates that no reducer is used.}
\label{tab:full_main}
\resizebox{\linewidth}{!}{
\begin{tabular}{llrrrr}
\toprule
Reducer & Metric & Cls. & OOD-Lag & OOD-TMoE & TL-TMoE \\
\midrule
\multirow{4}{*}{--}
    & MAD  & -0.041 & 0.210 & \textbf{0.619} & 0.309 \\
    & MMD  & 0.059 & 0.406 & 0.029 & -0.030 \\
    & OT   & \textbf{0.417} & 0.033 & 0.156 & 0.103 \\
    & WSD  & 0.010 & 0.273 & 0.604 & 0.353 \\
\midrule
\multirow{3}{*}{\rotatebox[origin=c]{90}{DBA}}
    & DTW  & 0.237 & \textbf{0.420} & 0.518 & 0.320 \\
    & ED   & 0.304 & 0.222 & 0.513 & \textbf{0.358} \\
    & LCSS & 0.127 & 0.202 & 0.087 & 0.012 \\
\midrule
\multirow{3}{*}{\rotatebox[origin=c]{90}{PCA}}
    & DTW  & -0.033 & 0.405 & 0.505 & 0.251 \\
    & ED   & 0.004 & 0.325 & 0.509 & 0.241 \\
    & LCSS & 0.006 & 0.272 & 0.068 & 0.051 \\
\bottomrule
\end{tabular}
}
\end{table}

\subsection{Analysis Layer}
\label{sec:analysis}

As shown in Fig.~\ref{fig:architecture}-(4), the Analysis Layer converts outputs from the Similarity Layer and Evaluation Layer into visual and quantitative analyses.
It provides heatmaps and network graphs for inspecting dataset relationships, and measures the alignment between similarity scores and downstream behavior using Pearson correlation~\cite{pearson1895vii} between similarity distances and task errors.
These analyses enable similarity metrics to be compared not only by the structures of their distance matrices, but also by their ability to explain classification and forecasting performance.

\section{Experiments}
We evaluate TSDS-Toolbox end-to-end across all layers.
The experiments use all supported similarity methods and downstream evaluation pipelines on 25 GluonTS datasets~\cite{alexandrov2020gluonts} spanning traffic, weather, electricity, exchange rates, tourism, and public health.
Each dataset is sampled with replacement to obtain 100 windows of length 100.
We apply z-score normalization and remove flat sequences before computing similarity scores.

We report Pearson correlation between dataset-distance scores and downstream task losses.
Table~\ref{tab:full_main} shows the correlation between dataset similarity and downstream task loss.
No similarity metric consistently achieves the highest correlation across all tasks, indicating that time-series dataset similarity is task-dependent.
In forecasting out-of-distribution (OOD) with Time-MoE~\cite{shi2025time}, \textbf{Match-and-Deform (MAD)}~\cite{painblanc2023match} achieves the strongest correlation, closely followed by \textbf{Wasserstein Distance (WSD)}~\cite{chen2025measuring}, suggesting that dataset-level metrics are particularly informative in this setting.
\textbf{Reducer-based dynamic time warping (DTW)}~\cite{sakoe1978dynamic} and \textbf{Euclidean Distance (ED)}~\cite{ding2008querying} remain competitive, whereas \textbf{Longest Common Subsequence (LCSS)}~\cite{vlachos2002discovering} and \textbf{Maximum Mean Discrepancy (MMD)}~\cite{gretton2012kernel} show weaker alignment.

\begin{figure}[t]
\centering
\includegraphics[width=0.95\linewidth]{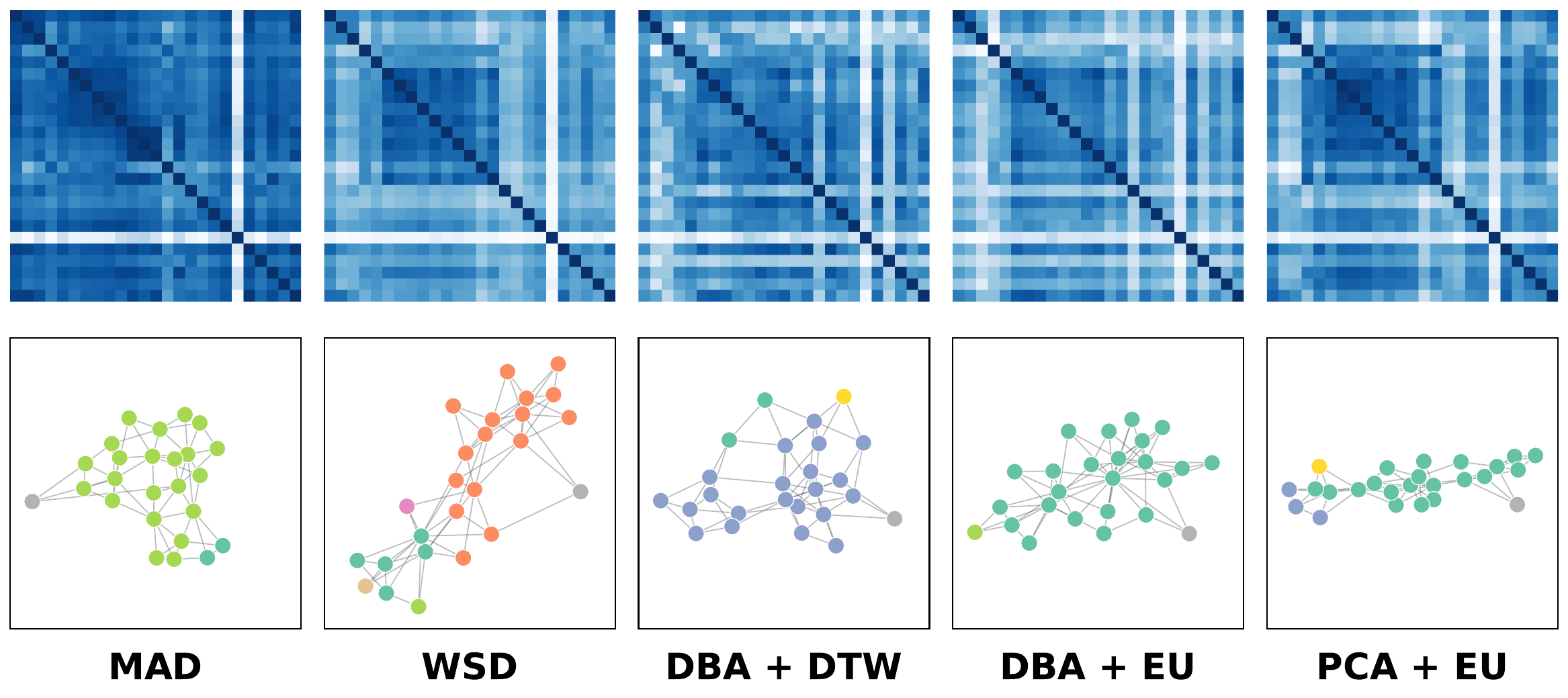}
\caption{Visualization results generated by TSDS-Toolbox.}
\label{fig:similarity_visualization}
\end{figure}

To further analyze reducer-based methods, Table~\ref{tab:reducer_comparison} compares \textbf{DTW Barycenter Averaging (DBA)}~\cite{petitjean2011global} and \textbf{Principal Component Analysis (PCA)}~\cite{pearson1901liii} as reducers.
DBA achieves stronger correlations in most settings, particularly for classification and Time-MoE~\cite{shi2025time} forecasting, whereas PCA remains competitive for Lag-Llama~\cite{rasul2023lag} OOD forecasting.
These results indicate that reducer effectiveness depends on both the downstream task and the underlying series-level distance.
Finally, Fig.~\ref{fig:similarity_visualization} presents heatmap and network-graph visualizations generated by TSDS-Toolbox, illustrating how different similarity metrics induce different dataset relationship structures over the same pool of datasets.

\begin{table}[t]
\centering
\small
\setlength{\tabcolsep}{6pt}
\caption{Reducer comparison across base distance metrics.}
\label{tab:reducer_comparison}

\begin{tabular}{llrrr}
\toprule
Task & Reducer & DTW & ED & LCSS \\
\midrule
Classification & DBA & \textbf{0.237} & \textbf{0.304} & \textbf{0.127} \\
               & PCA & -0.033 & 0.004 & 0.006 \\
\midrule
Forecasting OOD Lag-Llama & DBA & \textbf{0.420} & 0.222 & 0.202 \\
                           & PCA & 0.405 & \textbf{0.325} & \textbf{0.272} \\
\midrule
Forecasting OOD Time-MoE & DBA & \textbf{0.518} & \textbf{0.513} & \textbf{0.087} \\
                          & PCA & 0.505 & 0.509 & 0.068 \\
\midrule
Forecasting TL Time-MoE & DBA & \textbf{0.320} & \textbf{0.358} & 0.012 \\
                         & PCA & 0.251 & 0.241 & \textbf{0.051} \\
\bottomrule
\end{tabular}
\end{table}

\section{Conclusion}
In this paper, we presented TSDS-Toolbox, a unified toolbox for benchmarking time-series dataset similarity methods under standardized evaluation settings.
TSDS-Toolbox integrates similarity computation, downstream evaluation, and correlation-based analysis within a modular and extensible framework.
Our experiments show that no single similarity method consistently aligns with downstream performance across all tasks, highlighting the need for systematic evaluation rather than relying on a single metric.
By enabling reproducible benchmarking, TSDS-Toolbox supports fair comparison and practical source-dataset selection.
It also provides an extensible foundation for future research on time-series dataset similarity.

\bibliographystyle{plain}
\balance
\bibliography{references}

\end{document}